\documentclass{article}
\usepackage{times}
\usepackage{soul}
\usepackage{url}
\usepackage[hidelinks]{hyperref}
\usepackage[utf8]{inputenc}
\usepackage[small]{caption}
\usepackage{graphicx}
\usepackage{amsmath}
\usepackage{amsthm}
\usepackage{booktabs}
\usepackage{algorithm}
\usepackage{algorithmic}
\usepackage{xspace}

\usepackage[bm,microtype,nocitepackage]{pauli_new}

\usepackage[natbibapa]{apacite}

\usepackage{etoolbox}



\renewcommand{\algname}[1]{\textsc{#1}\xspace}
\newcommand{\grecondp}{\algname{GreConD+}}
\newcommand{\pandap}{\algname{PaNDa$^+$}}
\newcommand{\asso}{\algname{Asso}}
\newcommand{\bmad}{\algname{BMaD}}
\newcommand{\nassau}{\algname{nassau}}

\title{Recent Developments in Boolean Matrix Factorization\footnote{This
	technical report is the slightly extended version of a
	survey~\citep{miettinen20recent} which appeared at IJCAI'20.}}
\author{Pauli Miettinen\footnote{University of Eastern Finland, Kuopio, Finland.
	pauli.miettinen@uef.fi. Contact Author.} \and Stefan Neumann\footnote{KTH
		Royal Institute of Technology, Stockholm, Sweden.
			neum@kth.se. Work was done while
			the author was at University of Vienna, Vienna, Austria. The
			research leading to these results has received funding from the
			European Research Council under the European Community’s Seventh
			Framework Programme (FP7/2007-2013) / ERC grant agreement No.\
			340506, and the Doctoral Programme “Vienna Graduate School on
			Computational Optimization” which is funded by the Austrian Science
			Fund (FWF, project no. W1260-N35).}}

\begin{document}

\maketitle

\begin{abstract}
The goal of Boolean Matrix Factorization (BMF) is to approximate a given binary
matrix as the product of two low-rank \emph{binary} factor matrices, where the
product of the factor matrices is computed under the Boolean algebra. While the
problem is computationally hard, it is also attractive because the binary nature
of the factor matrices makes them highly interpretable.  In the last decade, BMF
has received a considerable amount of attention in the data mining and formal
concept analysis communities and, more recently, the machine learning and
the theory communities also started studying BMF. In this
survey, we give a concise summary of the efforts of all of these communities and
raise some open questions which in our opinion require further investigation.
\end{abstract}

\section{Introduction}
\label{sec:introduction}

Boolean matrix factorization (BMF) is a variant of the standard matrix
factorization problem in the Boolean semiring: given a binary matrix, the task
is to find two smaller \emph{binary} matrices so that their product, taken over the
Boolean semiring, is as close to the original matrix as possible. Because the
matrix product is not done over a field but over a semiring, many standard
matrix factorization techniques fail to work. Indeed, finding the best Boolean
factorization is computationally hard.

The computational hardness of the problem has not prevented people from studying
it. In psychometrics, some of the first algorithms appeared in the 1980's
(see \citet{belohlavek18new}). Even before that, mathematicians studying
combinatorics had studied the ``Boolean linear algebra''
\citep{kim82boolean,monson95survey}. More recently, \citet{miettinen06discrete}
introduced the problem to the data mining community, \citet{vaidya07role} to the access
control community, \citet{belohlavek10discovery} to formal concept analysts, and
\citet{broeck13complexity} to lifted inference community, to name but a few
examples. Even more recently, \citet{ravanbakhsh16boolean} studied the problem
in the framework of machine learning, increasing the interest of that community,
and \citet{chandran16parameterized,ban19ptas,fomin20approximation}
(re-)launched the interest in the problem in the theory community.

The various fields in which BMF is used are connected by the desire to
``keep the data binary''. This can be because the factorization is used as a
pre-processing step, and the subsequent methods require binary input, or because
binary matrices are more interpretable in the application domain. In
the latter case, Boolean algebra is often preferred over the field  $GF[2]$, as
the XOR-operation can be harder to interpret.

The large number of areas where BMF or related problems are studied has given
raise to many interesting approaches for solving this computationally hard
problem.  Unfortunately, it has also led to numerous re-inventions and repeated
effort.  Given the recent interest towards BMF in machine learning and AI
communities, we believe that a timely survey will help to guide the research to
novel directions and avoid re-inventions of older ideas.

To that end, we present a concise survey of the recent theoretical results on
the hardness of the problem in Section~\ref{sec:theory} and a survey on
different types of algorithms proposed for the problem in
Section~\ref{sec:algorithms}. We will provide some lessons we have learned when
developing and using some of these algorithms, as well as the main research
directions and open problems we find most interesting. Before going further,
however, we will provide the formal definitions of BMF in
Section~\ref{sec:definitions} and some applications thereof in
Section~\ref{sec:applications}.

\section{Definitions and Related Formulations}
\label{sec:definitions}

In this section we will explain the basic definitions for Boolean matrix
factorization (BMF) and related concepts. We will also cover the related formulations
of the problem using bipartite graphs and set systems. These formulations are
all equivalent, but they make connections to different problems more clear.

\subsection{Boolean Matrix Factorization}
\label{sec:basic-formulation}

The goal of matrix decompositions is to decompose (or factorize) a given input
matrix into two (or more) smaller \emph{factor matrices} whose product is close
to the original matrix. Thus, to define BMF, we first define the corresponding
product.

\begin{definition}
  \label{def:bprod}
  Given Boolean matrices $\mB\in\B^{m\times k}$ and $\mC\in\B^{k\times n}$, the
  \emph{Boolean product} of $\mB$ and $\mC$ is denoted
  $\mB\bprod\mC\in\B^{m\times n}$ and it is
  defined element-wise as
  \begin{equation}
    \label{eq:bprod}
    (\mB\bprod\mC)_{ij}
	= \bigvee_{\ell = 1}^k \mB_{i\ell}\mC_{\ell j}\; .
  \end{equation}
\end{definition}

The Boolean matrix product is like the standard matrix product of two binary-valued matrices, but the algebra is the Boolean semi-ring (the summation is defined as $1+1=1$). 

The \emph{Boolean rank} of matrix $\mA\in\B^{m\times n}$ is the least integer
$k$ such that there exists matrices $\mB\in\B^{m\times k}$ and
$\mC\in\B^{k\times n}$ with $\mA = \mB\bprod\mC$. In practice, the Boolean rank is rarely that interesting; like normal matrix rank, it is not robust to noise, and many real-world matrices have almost-full rank (though exceptions exist, see \citet{ene08fast}).

\begin{problem}
  \label{prob:bmf}
  Given a matrix $\mA\in\B^{m\times n}$ and integer $k$, the goal of \emph{Boolean matrix factorization} (BMF) is to find matrices $\mB\in\B^{m\times k}$ and $\mC\in\B^{k\times n}$ such that they minimize
  \begin{equation}
    \label{eq:bmf:def}
    \norm{\mA - \mB\bprod\mC}_F^2 = \sum_{i,j} \mA_{ij} \oplus (\mB\bprod\mC)_{ij} \; .
  \end{equation}
\end{problem}

In \eqref{eq:bmf:def}, subtraction, Frobenius norm $\norm{\cdot}_F$, and the
summation are taken over the standard algebra while $\oplus$ stands for the
element-wise XOR-operation. It might seem curious that the error is defined over
different algebra than the matrix product -- after all, the whole point of BMF
is that it uses the Boolean matrix product -- but it should be noted that the
Boolean algebra does not allow for subtraction. On the other hand, as all
elements of $\mA$ and $\mB\bprod\mC$ are binary, the (squared) Frobenius norm
simplifies into the Hamming distance between $\mA$ and $\mB\bprod\mC$, i.e.\ it is
simply the number of entries with $\mA_{ij}\neq(\mB\bprod\mC)_{ij}$.

\subsection{Covering by Bicliques}
\label{sec:covering-bicliques}

Any rectangular binary matrix $\mA\in\B^{m\times n}$ can be viewed as the
\emph{bi-adjacency matrix} of a bipartite graph $G = (U\cup V, E)$. The rows of
$\mA$ correspond to vertices in $U$, columns of $\mA$ correspond to vertices in
$V$, and $\mA_{ij} = 1$ if $\{i, j\}\in E$. Vice versa, any (unweighted,
undirected) bipartite graph $G$ can be identified with its bi-adjacency matrix
$\mA(G)$.

For bipartite graphs, BMF is equivalent to \emph{biclique cover}. 
\begin{problem}
  \label{prob:biclique-cover}
  Given a bipartite graph $G=(U\cup V, E)$ and an integer $k$, the goal of
  \emph{biclique cover} is to find $k$ \emph{bicliques} $C_1, \ldots, C_k$, such
  that they minimize
  \begin{equation}
    \label{eq:biclique-cover:def}
    \abs*[bigg]{E \oplus \bigcup_{\ell=1}^k E(C_{\ell})} \; .
  \end{equation}
\end{problem}
A \emph{biclique} is a complete bipartite subgraph
$C_\ell=(U_\ell \cup V_\ell, E(C_\ell))$, where $U_\ell\subseteq U$,
$V_\ell\subseteq V$ and $E(C_\ell) = U_\ell\times V_\ell$, i.e.\ $C_\ell$
contains edges between all pairs of vertices in $U_\ell$ and $V_\ell$.
Note that Problem~\ref{prob:biclique-cover} does not require that the edges of
$C_{\ell}$ are a subset of edges of $G$. Abusing the notation,  in
\eqref{eq:biclique-cover:def} the symbol $\oplus$ is denotes the
symmetric difference between sets.

The connection between Problems~\ref{prob:bmf} and \ref{prob:biclique-cover} can
observed as follows: Let $\vb_{\ell}$ be the $\ell$th column of $\mB$ and
$\vc^{\ell}$ be the $\ell$th row of $\mC$. Then we can write $\mB\bprod\mC$ as
the OR of outer products of $\vb_\ell$ and $\vc^\ell$, i.e.\ $\mB\bprod \mC =
\bigvee_{\ell=1}^k\vb_{\ell}\vc^{\ell}$. Each of these rank-one matrices (called
\emph{components}) $\vb_{\ell}\vc^{\ell}$ defines a rectangle of 1s, i.e.\
$(\vb_{\ell}\vc^{\ell})_{ij} = 1$ iff $(\vb_{\ell})_i = (\vc^{\ell})_j = 1$.
Hence, the bi-adjacency matrix corresponding to $\vb_{\ell}\vc^{\ell}$ defines a
biclique $C_\ell=(U_\ell, V_\ell)$ with $U_\ell=\{ i: (\vb_{\ell})_i=1\}$ and
$V_{\ell} = \{ j : (\vc^\ell)_j = 1 \}$. Then the bi-adjacency matrix of the
bipartite graph with edges $\bigcup_{\ell=1}^k E(C_{\ell})$ is
$\bigvee_{\ell = 1}^k \mB_{i\ell}\mC_{\ell j} = \mB\bprod\mC$.

\subsection{Set Systems}
\label{sec:set-systems}

A binary matrix $\mA\in\B^{m\times n}$ can also be identified as a \emph{set
system}: the columns correspond to $n$ items in universe $U$ and the rows are
incidence vectors of $m$ sets in a collection of sets
$\col{S} = \{ S_{i}\subseteq U : i=1,\ldots, m\}$. Then the following problem is
equivalent to BMF and will be called \emph{set basis}.
\begin{problem}
  \label{prob:set-basis}
  Given a set system $(U, \col{S})$ and an integer $k$, the goal of \emph{set basis}
  is to find a collection of $k$ sets
  $\col{C} = \{C_i \subseteq U : \ell = 1, \ldots, k\}$
  such that for every set $S_i\in\col{S}$ there exists a subcollection of
  $\col{B}_i\subseteq\col{C}$ that jointly minimize
  \begin{equation}
    \label{eq:set-basis:def}
    \sum_{i=1}^m \abs*[bigg]{S_i \oplus \bigcup_{C\in\col{B}_i}C}\; .
  \end{equation}
\end{problem}

In this formulation, the rows of $\mA$ correspond to the sets $S_i$ with
$S_i=\{ j : \mA_{ij}=1 \}$. The matrix $\mC$ encodes the incidence
vectors of the sets $C_\ell\in\col{C}$, i.e.\ $\mC_{\ell i}=1$ iff $i \in C_\ell$.
The choice of which sets to include in which subcollection $\col{B}_i$ is
given by the rows of $\mB$, i.e.\ $\mB_{i \ell} = 1$ iff $C_{\ell}\in
\col{B}_i$. Thus, Problem~\ref{prob:set-basis} is equivalent to
Problem~\ref{prob:bmf}.

In \emph{tiling}~\citep{geerts04tiling}, we are given a transactional database,
that is, a collection of sets (transactions) over items and the goal is to cover
the transactions with itemsets. In standard tiling, no \emph{over-covering} is
allowed. That is, using notation from Problem~\ref{prob:set-basis}, it is
required that $\bigcup_{C\in\col{B}_i}C \subseteq S_i$. In noisy tiling, which
is equivalent to BMF, this restriction is lifted, allowing \emph{over-covering}
(i.e.\ if an element $j\in \bigcup_{C\in\col{B}_i}C$ is not in $S_i$, then
 $(\mB\bprod\mC)_{ij}=1$ but $\mA_{ij} = 0$).

\section{Applications}
\label{sec:applications}

Let us now discuss applications of BMF.
\Citet{wicker12multi-label} use BMF to pre-process multi-label classification
data. They decompose the binary label matrix and use the columns of the left
factor matrix $\mB$ as sort-of ``super-labels.'' For a recent review of this and other approaches for dimensionality reduction for multi-label classification, see \citet{siblini19review}.

Somewhat similarly, \citet{broeck13complexity} use BMF to speed up lifted
inference in Boolean data. They show that the inference is efficient if the data
has low Boolean rank, and to transform the data, they compute
low-rank Boolean decomposition before doing the inference. 

\Citet{vaidya07role} propose using BMF
for \emph{role mining}. In role mining, we are given a binary matrix that
represents users and permissions, and the goal is to assign users to (possibly
overlapping) roles and to give these roles necessary permissions. Hence, if
$\mA$ is the user--permission matrix, $\mB$ assigns users to $k$ different roles
and $\mC$ assigns the permissions to these roles. \Citet{ene08fast} showed that
many real-world matrices in this domain actually have low Boolean rank.

BMF has many more applications and due to lack of space, we can only name a few.
In bioinformatics, BMF is used for studying transcriptomic
data~\citep{liang20bem} and for identifying functional interactions in brain
networks~\citep{haddad18identifying}. BMF was also used for adding missing
attribute information to products in the Amazon
catalog~\citep{rukat17interpretable}. Further applications include approximate
logic synthesis~\citep{hashemi18blasys} and discovery of patterns in network
data~\citep{kocayusufoglu18summarizing}.

\section{Theoretical Results}
\label{sec:theory}
In this section we summarize computational lower bounds and upper bounds for
computing BMF.

\subsection{Computational Complexity}
\label{sec:complexity}
Let us review the computational complexity of 
Problem~\ref{prob:bmf}.

\Citet{orlin1977contentment} showed that, given
a bipartite graph $G$ and a parameter $k$, it is \NP-complete to decide whether
the edges of $G$ can be covered with $k$ bicliques. Via the discussion from
Section~\ref{sec:covering-bicliques} this implies that it is \NP-complete to
decide whether there exist Boolean matrices $\mB$ and $\mC$ of rank $k$ such
that $\norm{\mA - \mB\bprod\mC}_F^2 = 0$. Hence, Problem~\ref{prob:bmf}
is \NP-complete.

\Citet{chandran16parameterized} showed that under a standard assumption in
computational complexity theory, one cannot distinguish between the case when the
objective function in Problem~\ref{prob:bmf} is $0$ or at least $1$ in time
$2^{2^{o(k)}} (mn)^{O(1)}$. This implies that either the running time of an
algorithm must be $2^{2^{\Omega(k)}}$ or it must be $(mn)^{\omega(1)}$, i.e.\  
the algorithm's running time is either doubly exponential in $k$ or
superpolynomial in the size of the input matrix.

In fact, the above results hold for any $\alpha$-approximate solution. Here, we
say that matrices $\mB'\in \{0,1\}^{m\times k}$ and $\mC'\in\{0,1\}^{k\times n}$
provide an \emph{$\alpha$-approximation} of the optimal objective function value
if
\begin{align*}
	\norm*{\mA - \mB'\bprod\mC'}_F^2
  \leq \alpha \cdot \min_{
  \mB,\mC}
  \norm*{\mA - \mB\bprod\mC}_F^2\; .
\end{align*}
Observe that any $\alpha$-approximation must be able to distinguish between
the case that $\min_{ \mB, \mC } \norm{\mA - \mB\bprod\mC}_F^2=0$ or
$\min_{ \mB, \mC } \norm{\mA - \mB\bprod\mC}_F^2>0$. Hence, the above results
imply that for any $\alpha\geq 1$, computing an $\alpha$-approximation for
Problem~\ref{prob:bmf} is \NP-complete and cannot be solved in time
$2^{2^{o(k)}} (mn)^{O(1)}$ under a standard complexity assumption.

\subsection{Theoretical Algorithms}
\label{sec:theory-algos}
Let us now look at algorithms with theoretical guarantees.

\Citet{gramm08data} and \citet{fomin20parameterized} showed that
an \emph{exact} solution for BMF can be computed in time $2^{2^{O(k)}}
(mn)^{O(1)}$, thus matching the above lower bound.

Furthermore, \citet{fomin20approximation} provided an algorithm which computes a
$(1+\varepsilon)$-approximate BMF in time $2^{2^{O(k)}/\varepsilon^2 \cdot \lg^2(1/\varepsilon)} \cdot mn$.
Note that when $k$ and $\varepsilon$ are constants, then the running time of
this algorithm becomes $O(mn)$, i.e.\ the algorithm runs in time linear in the
size of the $m \times n$ input matrix $\mA$.  This
is unlike the exact algorithms, where the exponent of the $mn$-term is strictly
larger than $1$, i.e.\ they require superlinear running time even when $k$ is a
fixed constant.  \citet{ban19ptas} obtained a similar algorithm with slightly
worse running time, but their algorithm extends to any finite field.

\Citet{bhattacharya19streaming} extended ideas of \citet{fomin20approximation}
and \citet{ban19ptas} to obtain a 4-pass streaming algorithm which computes a
$(1+\varepsilon)$-approximate BMF.  Their algorithm never stores more than
$2^{\tilde{O}(2^k /\varepsilon^2)} \cdot (\lg n)^{2k}$ rows of the matrix and it
has running time $2^{\tilde{O}(2^k /\varepsilon^2)} \cdot (\lg n)^{2k} \cdot mn$.
Note that when $k$ is a fixed constant, then the algorithm only stores a
polylogarithmic number of rows.

\subsection{Discussion}
From a theoretical perspective, BMF is relatively well-understood.  Indeed, note
that the lower bound results by \citet{chandran16parameterized} and the running
times of the algorithms by \citet{fomin20approximation} and \citet{ban19ptas}
match up to lower order terms. Hence, there is no hope in removing the
$2^{2^{O(k)}}$-terms in the running times of the algorithms.

Unfortunately, all of the above algorithms are rather impractical.  First of all,
the $2^{2^{O(k)}}$-term in the running time grows extremely fast (even for
$k=7$, $2^{2^7} > 3\cdot 10^{38}$).  Secondly, the algorithms have in common
that they use several exhaustive enumeration procedures which are extremely slow
in practice.

There are two main open questions in this area. First, can one improve upon the
results by \citet{bhattacharya19streaming} by providing a streaming algorithm
which performs less than $4$ passes and returns a BMF.  Second, it would be
interesting whether there are practical algorithms for which one can obtain
similar running time and approximation guarantees to the ones mentioned above.

\section{Algorithms}
\label{sec:algorithms}

In this section, we discuss several practical algorithms for computing BMF.
From a high-level point of view, the algorithms can be placed into two categories:
Those based on combinatorial optimization and those based on continuous
optimization. We will now discuss each type of these algorithms and conclude the
section with some open questions.

\subsection{Combinatorial Optimization}
\label{sec:combinatorial}

Some of the earliest algorithms for BMF come from psychometrics community. One example is the 8M algorithm in the BMDP statistical software (see \citet{belohlavek18new} for a detailed description). The 8M algorithm follows a simple local update process: it starts by building initial matrices $\mB$ and $\mC$, and then iteratively replaces the initial columns of $\mB$ and rows of $\mC$ with (locally) better ones.

The iterative update approach is also used by \citet{belohlavek18new} in their
\grecondp algorithm. \grecondp takes carefully into account the
anti-monotonicity of Boolean algebra: if one component $\vb_{\ell}\vc^{\ell}$
introduces a $1$ in location $(i,j)$ where $\mA_{ij} = 0$, this error is
irrecoverable by the other components. Hence, \grecondp starts by generating a
decomposition that does not over-cover (i.e.\ they generate a tiling) and then
the components are iteratively updated to commit over-covering when that
decreases the error. After one component is updated, other components are
examined if they can be edited -- or removed entirely -- to improve the error.

The \grecondp algorithm belongs to a collection of BMF algorithms based on
\emph{formal concept analysis} \citep{ganter99formal} (see also \citet{belohlavek10discovery,belohlavek15from-below}). A formal concept, in the
BMF framework, is a pair of row and column index sets $(I, J)$ such that
$\mA_{ij} = 1$ for all $i\in I$ and all $j\in J$ and such that no new index can
be added to neither $I$ or $J$. Such a pair of indices corresponds to components
$\vb_{\ell}\vc^{\ell}$ that do not over-cover (they also correspond to
		\emph{closed itemsets} in frequent itemset mining). Most formal-concept
based algorithms find non-overcovering factorizations as a
consequence.

The tiling approach also motivated the \pandap
algorithm~\citep{lucchese13unifying}. However, instead of finding strictly
non-overcovering initial components, as \grecondp does, \pandap finds
\emph{dense} initial components, that is, initial components that overcover only
strictly limited amount of elements. In the second phase of the algorithm, these
dense cores are extended, again provided that the extension does not increase
the error. An important feature of \pandap is that the number of components,
$k$, is not predefined. Rather, \pandap selects the number of components
automatically, based on user-selected criteria for balancing the decrease in
error and increase in the number of components.

Yet another approach is presented by the \asso
algorithm~\citep{miettinen08discrete}. Unlike the previously-described
algorithms, \asso starts by generating a large collection of candidate columns
for $\mB$. There will be a candidate column for every row $\va^i$ of $\mA$. A
candidate column is generated by comparing the association confidence from
$\va^i$ to $\va^j$, for any $j=1,\ldots, m$, to a user-supplied threshold $\tau$.
The candidate column will have a $1$ in the $j$th entry if the confidence,
computed as $\va^i(\va^j)^T/(\va^i(\va^i)^T)$, is above the threshold $\tau$. 

\asso then greedily selects the candidates one-by-one to be added to $\mB$ by
computing how much error each candidate would reduce. In order to compute the
error, \asso computes for each column $\va$ of $\mA$ and for each candidate
column $\vd$, how much the reconstruction error of $\va$ would change if $\vd$
would be used to cover the column, that is, it computes
$\norm{\va - \mB\bprod\vc}_F^2 - \norm{\va - [\mB\; \vd]\bprod\left[\begin{smallmatrix}\vc\\1\end{smallmatrix}\right] }_F^2$,
where $\mB$ is the current left factor matrix 
and $\vc$ is the column of the right factor matrix corresponding to $\va$. The
candidate is used in every column of $\mA$ where it reduces the error, and the
candidate that reduces the error the most over all columns of $\mA$ is selected.
The corresponding row of $\mC$ is built based on the greedy computation: if the
selected $\vd$ reduces the error on column $j$, the corresponding column in
$\mC$ will be set to $1$.

As mentioned, \pandap uses some measure to balance the reduction in error and the size of the factorization. There are different ways to do this, but one of the popular and principled ways for solving the \emph{model order selection} problem is to use the \emph{minimum description length} (MDL) principle~\citep{rissanen78modeling}. \Citet{miettinen14mdl4bmf} proposed to use two-part MDL for selecting the number of components in BMF. Two-part MDL computes the encoding length of the data $L(\mA)$ as the length of the model (or hypothesis) $L(\mathcal{M})$ plus the length of the data given the model $L(\mA\mid \mathcal{M})$. In case of BMF, the model consists of the factor matrices, while the data given the model is the error the factorization causes. That is,
\begin{equation}
  \label{eq:mdl4bmf}
  L(\mA) = L(\mB,\mC) + L\bigl(\mA\oplus(\mB\bprod\mC)\bigr)\; .
\end{equation}
The length of encoding a binary matrix can be computed using standard methods for binary string encoding. \Citet{miettinen14mdl4bmf} propose the use of so-called \emph{data-to-model} encodings and differentiating over- and under-covering noise.

\Citet{miettinen14mdl4bmf} use a variation of \asso to find the decomposition
that minimizes \eqref{eq:mdl4bmf}. Also \pandap can be used to find such
decomposition. \Citet{karaev15getting} also try to minimize \eqref{eq:mdl4bmf}
and their algorithm, called
\nassau, shares similarities with many of the above algorithms: it starts by
finding a set of seed columns (akin to candidates in \asso). Unlike
\asso, though, \nassau uses random walks in the bipartite graph corresponding to
$\mA$ to find the seeds. In short, they consider the probability of a
random walk starting from a vertex $u\in U$ to be in another vertex
$u'\in U$. The seed column corresponding to vertex $u\in U$ has 1-entries in the
rows corresponding to the vertices $u'$ where the random walk would be with a
sufficiently high probability. The seed columns are then turned into rank-1
components using a similar greedy strategy as used by \asso. After a new rank-1
component is added, the existing ones are iteratively updated, similarly to 8M,
potentially dropping some if they become redundant.

Many of the above algorithms first build some initial solution or candidate set,
and then refine it. The \bmad framework \citep{tyukin14bmad} formalizes this
process and allows the user to define the approach for candidate generation, the
approach for selecting the candidates, and the approach for building the
factorization from the candidates.

\subsection{Continuous Optimization}
\label{sec:learning}
Next, let us look at continuous optimization algorithms.

\Citet{hess17primping} propose to use a proximal alternating linearized
minimization technique for finding the MDL-optimizing BMF. The idea is to
replace BMF with a surrogate continuous optimization problem with a two-part goal:
minimizing the error and making the factor matrices binary. This approach allows
alternating between gradient descent optimization of the factor matrices and
computing the proximal function. \Citet{hess18trustworthy} use a similar
technique for finding Boolean decompositions that limit the \emph{false
discovery rate}, i.e.\ the probability that the found patterns arise from noise.

Next, we discuss a sequence of works which try to find factor matrices $\mB$ and
$\mC$ with maximum likelihood under the given input matrix $\mA$. This includes works of 
\citet{ravanbakhsh16boolean,rukat17bayesian,rukat18tensormachine,liang19noisy,frolov14expectation}.
In order to compute the likelihoods of the factor matrices, they make
some prior assumptions on the data-generating process.  We
will only discuss the model of \citet{ravanbakhsh16boolean} in detail.

From a high-level point of view, \citet{ravanbakhsh16boolean} assume that the
input matrix $\mA$ for Problem~\ref{prob:bmf} is generated as follows. First,
two ground-truth factor matrices $\mB\in\{0,1\}^{m\times k}$ and
$\mC\in\{0,1\}^{k\times n}$ are generated randomly according to a certain
distribution. This gives a matrix $\mZ=\mB\bprod\mC$ of Boolean rank
$k$. Now the matrix $\mA$ is generated from $\mZ$ by setting $\mA=\mZ$ and then
flipping the bit in each entry $\mA_{ij}$ according to some distribution.

Let us now look consider this model in more detail.
For all entries of the factor matrices
$\mB\in\{0,1\}^{m\times k}$ and $\mC\in\{0,1\}^{k\times n}$ there exist priors
$p_{ir}^{\mB}(\cdot)$ and $p_{rj}^{\mC}(\cdot)$ for all $i=1,\dots,m$,
$r=1,\dots,k$ and $j=1,\dots,n$. In particular, it is assumed that $\mB_{ir}=1$
($\mC_{rj}=1$) with probability $p_{ir}^{\mB}(\mB_{ir})$ ($p_{rj}^{\mC}(\mC_{rj})$)
and that for both factor matrices the prior takes the following separable
product form:
\begin{align*}
	p^{\mB}(\mB) = \prod_{i,r} p_{ir}^{\mB}(\mB_{ir})\; ,
	\hspace{0.5cm}
	p^{\mC}(\mC) = \prod_{r,j} p_{rj}^{\mC}(\mC_{rj})\; .
\end{align*}
Next, given $\mB$ and $\mC$ as above, consider the product matrix
$\mZ=\mB\bprod\mC\in\{0,1\}^{m\times n}$ and observe that $\mZ$ only depends on
the randomness of the entries in $\mB$ and $\mC$. It is assumed that for all
$i=1,\dots,m$ and $j=1,\dots,n$ there exists another prior
$p^{\mA}_{ij}(\mA_{ij} \mid \mZ_{ij})$, i.e.\ the probability of flipping the bit
in entry $\mA_{ij}$ can depend on $i$, $j$ and $\mZ_{ij}$. It is further assumed
that for the matrix $\mA$ it holds that
\begin{align*}
	p^{\mA}(\mA \mid \mZ)
	= \prod_{i,j} p_{ij}^{\mA}(\mA_{ij} \mid \mZ_{ij})\; .
\end{align*}
Recall that in Problem~\ref{prob:bmf} the input is the matrix $\mA$ and we
are actually interested in computing the (unknown) factor matrices $\mB$ and
$\mC$. The approach of \citet{ravanbakhsh16boolean} is to solve the maximum a
posteriori inference problem
\begin{align*}
	\argmax_{\mB,\mC} p(\mB, \mC \mid \mA)\; .
\end{align*}
To solve the above inference problem, they introduce a graphical model for the
posterior and use belief propagation to find the matrices $\mB$ and $\mC$ which
maximize the posterior.

\Citet{frolov14expectation} and \citet{liang19noisy} used a similar generative
model and expectation maximization instead of belief propagation.
\Citet{rukat17bayesian} used a Metropolised Gibbs sampler for the posterior
inference. Later, \citet{rukat18tensormachine} extended their algorithm for Boolean tensors.

\subsection{Open Problems and Lessons Learned}
\label{sec:algos-discussion}

The algorithms based on combinatorial optimization use various heuristics and
iterative update approaches to avoid the problem that their optimization
landscape is not at all smooth. Indeed, most iterative update algorithms will
converge in just few rounds (see, e.g.\ the experiments by
\citet{miettinen09matrix}). No currently-published algorithm combines all of the
approaches (many candidates, preference for under-covering, iterative local
updates, etc.). Whether a combination of these approaches would improve the
results will remain as a topic of further research, although our experience
indicates that combining multiple approaches has diminishing returns in most
cases.

At the same time it should also be noticed that while it is easy to build
examples where greedy heuristics will fail spectacularly, they often tend to
work very well in real-world applications. In case of BMF, the experiments
by \citet{miettinen09matrix} seem to confirm this at least when it comes to the
problem of selecting the candidates to build the factorization.

When the data is very sparse, some algorithms based on continuous
optimization outperform all combinatorial algorithms we are aware of.
Specifically, the experiments by \citet{ravanbakhsh16boolean} show that on
synthetic data, their algorithm recovers planted structures close to what is
possible (they show that the reconstruction error of
their algorithm is close to an information theoretic lower bound by
\citet{davenport14bit}).  This is also supported by the experiments by
\citet{neumann18bipartite}.

A major open question is to make existing algorithms more
scalable. Indeed, existing algorithms do not easily scale to matrices with
tens or hundreds of thousands rows and columns. For the combinatorial optimization
algorithms, a major bottleneck is that they need to generate a lot of candidate
factor matrices and, hence, need to cover the input matrix many times. For the
continuous optimization algorithms, a major bottleneck is that usually they
maintain dense real-valued versions of the factor matrices; this makes them slow
and memory-intensive.
However, these methods may benefit from the recent GPU development and
methods such as stochastic gradient descent;
indeed, such techniques are used by \citet{hess17primping}.
In practice, the scalability problem is often bypassed by
pruning all rows and columns from the input matrix which have less than a
user-specified threshold of non-zero entries. 

None of the above algorithms comes with provable guarantees to return a solution
within a certain approximation ratio. While some of them will converge to an
optimal solution, it is not clear how many iterations will be necessary.

\section{Related Work}
\label{sec:related}

BMF has been extended (and constrained) in various forms to work better in different applications.

\Citet{tatti19boolean} aim at finding BMF where the rows and columns of the data
and factor matrices can be permuted such that the $1$s in the columns of $\mB$
and in the rows of $\mC$ all appear consecutively. This is used for sorting
vertices when graphs are drawn as edge bundles.

Many datasets are multi-view: they present different data about the same
entities. Sometimes such data can be expressed as two (or more) matrices, and
they can be jointly decomposed. \Citet{miettinen12finding} solves \emph{joint BMF}: given $\mA$ and $\mB$ whose rows represent the same entities, find
	$\mW$, $\mU$, $\mV$, $\mH$, and $\mL$ such that $\mA \approx [\mW\;
	\mU]\bprod\mH$ and $\mB\approx[\mW\; \mV]\bprod\mL$. Similarly,
	\citet{hess17c-salt} find decompositions of type
	$\mA\approx(\mW\lor\mU)\bprod\mH$ and $\mB\approx(\mW\lor\mV)\bprod\mL$
	(note that the dimensions of the matrices are not the same in these two definitions).

One of the most actively studied extension of BMF is Boolean tensor
decompositions~(see, e.g.~\citet{miettinen11boolean,erdos13walknmerge,rukat18tensormachine}). Briefly, tensors extend the matrix (2-way) data to multi-way data. For example, a
Boolean rank-$r$ CP decomposition of a three-way tensor $\tA$ would decompose it
into three factor matrices, $\mB$, $\mC$, and $\mD$, minimizing
\[
\sum_{i,j,k}\abs*[Big]{\tA_{ijk} - \bigvee_{\ell=1}^r\mB_{i\ell}\mC_{j\ell}\mD_{k\ell}}\; .
\]

While BMF can be seen as covering a bipartite graph with (potentially
overlapping) bicliques, many
algorithms have been proposed for the problem of covering a bipartite graph with
\emph{disjoint} bicliques. These problems are often studied under names such as
\emph{biclustering}~\citep{lim15convex},
\emph{co-clustering}~\citep{dhillon01coclustering} or
\emph{bipartite graph partitioning}~\citep{zha01bipartite}.
\Citet{kumar19faster} provide a constant factor approximation algorithm for
covering a bipartite graph with disjoint bicliques with running time
$2^{O(k^2)} (mn)^{O(1)}$.  This is an exponential improvement in $k$ over the
results mentioned in Section~\ref{sec:complexity}, but it requires the bicliques
to be disjoint.

Covering \emph{random} bipartite graphs with disjoint bicliques has recently
received some
attention~\citep{lim15convex,xu14jointly,zhou18optimal,zhou19analysis,ndaoud19improved}.
In these \emph{bipartite stochastic block
models} one assumes that the input bipartite graph was randomly generated
based on some hidden bicliques. The goal is to state
conditions under which the hidden bicliques can be recovered when the input
only consists of the random bipartite graph.  These random graph models are
similar to the data-generating models mentioned in Section~\ref{sec:learning}
(assumptions on the hidden bicliques correspond to priors on factor
 matrices).  \Citet{neumann18bipartite} showed that in such random bipartite
graphs a simple two-step algorithm can recover even very tiny clusters, which is
not possible in general graphs.

\section{Future Directions and Conclusion}
\label{sec:future}

In this survey, we have summarized the current state of the art for BMF
algorithms including several recent developments based on results from the
machine learning and theoretical computer science community.

From a practical point of view, perhaps the most important open problem is to
make the existing algorithms more scalable. A first step in this direction was
made by~\citet{neumann20biclustering}, who presented a streaming algorithm for
BMF.

From a theoretical point of view, the lower bounds from
Section~\ref{sec:complexity} essentially rule out any \emph{practical}
algorithms which provably approximate BMF. Nonetheless,
we have discussed in Section~\ref{sec:algos-discussion} that there are algorithms
which appear to recover very high quality results in practice.
Hence, we believe that one should try to analyze BMF algorithms not using
the current worst-case analysis approach, but using other complexity measures.
In our opinion, one promising approach would be to generalize the results for
bipartite stochastic block models we mentioned in Section~\ref{sec:related} from
disjoint biclique covers to overlapping biclique covers. This would imply
practical algorithms for BMF with provable quality guarantees.  Since some of
the bipartite stochastic block model algorithms work well in practice and their
running times do not carry large overheads in some parameters as the ones from
Section~\ref{sec:theory-algos}, we believe that there is hope to obtain
such results.

\bibliographystyle{apa}
\bibliography{main}

\end{document}